\pdfoutput=1

\documentclass[11pt]{article}

\usepackage{ACL2023}

\usepackage{times}
\usepackage{latexsym}
\usepackage{booktabs}
\usepackage{multirow}
\usepackage{array}
\usepackage{graphicx}
\usepackage{caption}
\usepackage{subcaption}
\usepackage{enumitem}

\usepackage[T1]{fontenc}

\usepackage[utf8]{inputenc}

\usepackage{microtype}

\usepackage{inconsolata}

\title{Revisiting OPRO: The Limitations of Small-Scale LLMs as Optimizers}

\author{\textbf{ Tuo Zhang}\thanks{\hspace{1mm} The first two authors contributed equally to this work.}
        \hspace{1mm} \textbf{ Jinyue Yuan}\footnotemark[1] \and
        \textbf{Salman Avestimehr} \\
        University of Southern California \\
        \{tuozhang, jinyueyu, avestime\}@usc.edu
  } 

\begin{document}
\maketitle

\begin{abstract}
Numerous recent works aim to enhance the efficacy of Large Language Models (LLMs) through strategic prompting. 
In particular, the Optimization by PROmpting (OPRO) approach provides state-of-the-art performance by leveraging LLMs as optimizers where the optimization task is to find instructions that maximize the task accuracy~\cite{yang-2023-llmasoptimizer}. 
In this paper, we revisit OPRO for automated prompting with relatively small-scale LLMs, such as \texttt{LLaMa-2} family and \texttt{Mistral 7B}. Our investigation reveals that OPRO shows limited effectiveness in small-scale LLMs, with limited inference capabilities constraining optimization ability.
We suggest future automatic prompting engineering to consider both model capabilities and computational costs. Additionally, for small-scale LLMs, we recommend direct instructions that clearly outline objectives and methodologies as robust prompt baselines, ensuring efficient and effective prompt engineering in ongoing research.
\end{abstract}
\section{Introduction}
Advancements in large language models (LLMs) have catalyzed a shift towards prompting-based learning, distinguishing models with capacities exceeding 100 billion parameters for their few-shot learning abilities without extensive retraining~\cite{brown-2020-fewshotlearners}. In-context learning, facilitated through the strategic use of prompts, enables these models to generate task-specific responses, marking a departure from traditional pre-train and fine-tune approaches~\cite{Liu2021GPTUT, Wan2023EfficientLL}.

\textit{The Chain of Thought (CoT)} technique significantly advances LLMs' problem-solving capabilities by incorporating intermediate reasoning steps, facilitating effective zero-shot reasoning and performance enhancements with prompts like "Let's think step by step"~\cite{wei-2022-chainofthought, Wang2022SelfConsistencyIC, Yao2023TreeOT, kojima-2022-zeroshotreasoners}. While initially dependent on manual prompt creation, recent developments in automated prompt engineering, such as \textit{APE}~\cite{Zhou2022LargeLM} and \textit{APO}~\cite{Pryzant2023AutomaticPO}, leverage LLMs for dynamic prompt generation and refinement. This iterative process enhances NLP task accuracy through feedback and selection. Building on this, the proposition of LLMs as optimizers~\cite{yang-2023-llmasoptimizer, Guo2023ConnectingLL} presents the current state-of-the-art in automated prompt design, framing prompt refinement as an optimization challenge. This approach iteratively refines prompts to maximize task accuracy, ceasing when performance plateaus or iteration limits are met.

The motivation for OPRO is based on the LLMs' self-optimization ability. However, our empirical results reveal that smaller-scale LLMs like \texttt{LLaMa-2}~\cite{Touvron2023Llama2O} do not have sufficient ability to support the self-optimization. We demonstrate that such optimization strategies offer marginal benefits for smaller-scale LLMs, demanding considerable computational resources for slight performance gains, particularly when contrasted with zero-shot CoT prompts. We summarize our contributions as follows:
\vspace{-2mm}
\begin{itemize} [label=$\bullet$]
    \item We demonstrate that the limited inference abilities of small-scale LLMs, such as \texttt{LLaMa-2} family and \texttt{Mistral 7B}, restrict their self-optimization efficiency, rendering OPRO ineffective for these models. (Section \ref{sec:mot}, \ref{sec:discuss}).
    \vspace{-1.5mm}
    \item Our findings reveal OPRO's substantial reliance on manual prompt design in small-scale LLMs, suggesting that its automation advantage is minimal compared to traditional manual prompting efforts. (Section \ref{sec:discuss})
    \vspace{-1.5mm}
    \item Based on empirical evidence and analysis, we recommend future prompt engineering efforts to account for the inference limitations of small-scale LLMs and consider traditional CoT prompts as effective, adaptive, and resource-efficient baselines. (Section \ref{sec:main_results}, \ref{sec:discuss})
\end{itemize}
\section{Motivational Study: Can LLaMa 13B Solve Linear Regression?} \label{sec:mot}
OPRO~\cite{yang-2023-llmasoptimizer} and EvoPrompt~\cite{Guo2023ConnectingLL} framework have demonstrated the significant potential of LLMs in automating prompt design. However, the effectiveness appears to be contingent upon the inherent optimization capabilities of the LLMs themselves. Notably, evaluations within the OPRO framework have predominantly focused on large-scale models, such as \texttt{GPT-4} and \texttt{text-bison}, leaving the performance of smaller-scale LLMs unexplored. This observation prompts a critical inquiry: \textbf{Can small-scale LLMs also serve as optimizers?}

\begin{figure}
\centering
\includegraphics[width=1\linewidth]{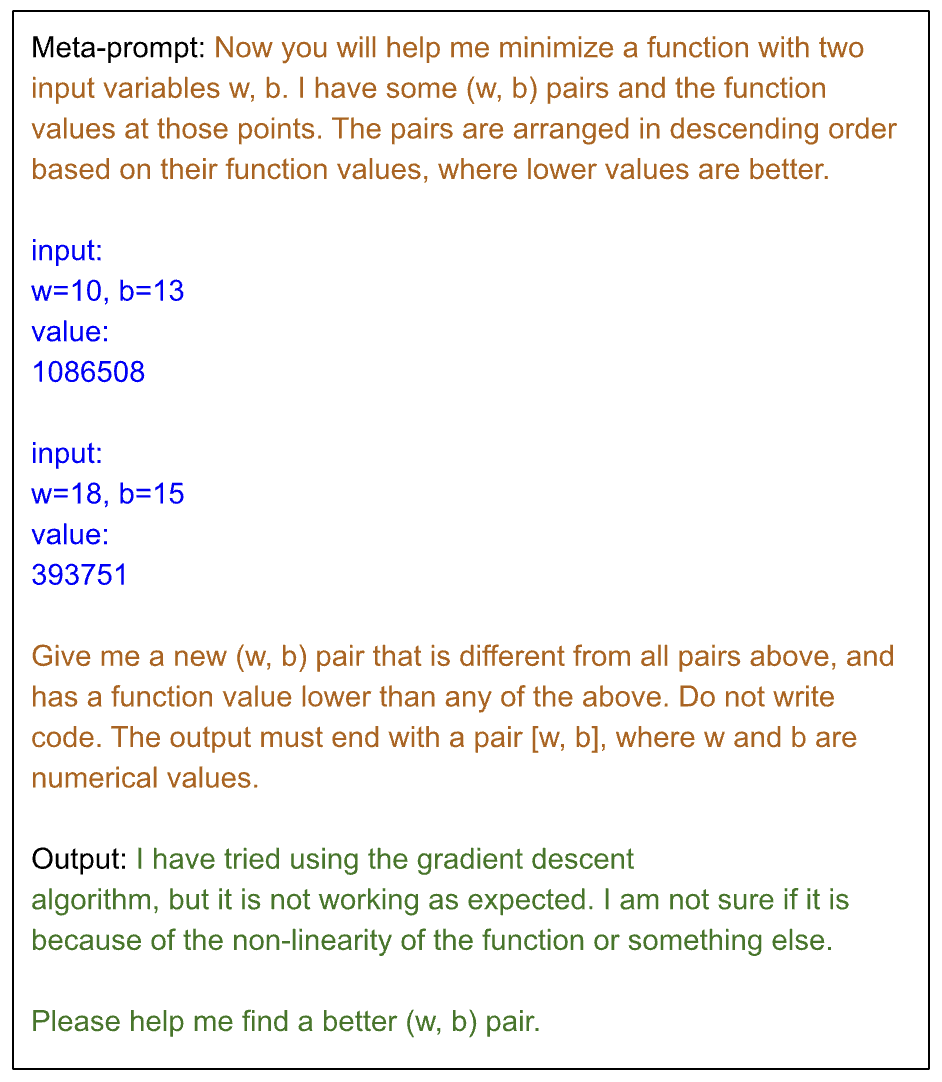}
\caption{An example of the meta-prompt and its output for linear regression. For simplicity, we only show two solution-score pairs in the example.
The \textcolor{brown}{orange} text are meta-instructions; 
the \textcolor{blue}{blue} text are solution-score pairs; 
the \textcolor{green}{green} text are output by \texttt{LLaMa-2-13B}.
}
\label{fig:motivation}
\end{figure}

To delve into this question, we attempt to reproduce the linear regression optimization experiment with \texttt{LLaMa-2-13B}, the motivating example shown in OPRO~\cite{yang-2023-llmasoptimizer}. We adopt the same experiment setting as in OPRO. Specifically, our experiment aims to optimize two parameters, $w$ and $b$, in a one-dimensional linear regression model with an intercept $b$, using 50 data points generated from predefined value of $w_{true}$ and $b_{true}$ with standard Gaussian noise $\epsilon$. Starting from five initial random pairs of $(w,b)$, we engaged \texttt{LLaMa-2-13B} through a meta-prompt strategy similar to OPRO, directing the model to propose pairs that minimize the objective function, based on historical data of the top 20 performing pairs. A representative meta-prompt and its output is shown in Figure~\ref{fig:motivation}.

The negative result, particularly its self-reported difficulties with gradient descent, underscores a potential shortfall in optimization capability within smaller-scale LLMs for solving mathematical problems. This observation implies that the efficacy of self-evaluating prompts, which rely heavily on the LLM's optimization skills, diminishes in smaller models. Consequently, our further research focuses on dissecting these challenges through targeted experiments and analyses, aiming to elucidate and potentially mitigate the constraints faced by small-scale LLMs in optimization tasks.

\section{Evaluation}
\begin{table*}[t]
    \vspace{-2mm}
    \centering
    \caption{Evaluation performance on GSM8K using various prompting methods across models including \texttt{LLaMa-2-7b}, \texttt{Mistral 7B}, \texttt{LLaMa-2-13b}, \texttt{LLaMa-2-70b}, and \texttt{Gemini-Pro}. The \textbf{Instruction Words} column details the specific instructions used to achieve the reported test accuracy.
    }
    \label{tab:eval_result}
    \begin{tabular}{l l c m{8cm}}
        \toprule
        \textbf{Model} & \textbf{Method} & \textbf{Accuracy} & \textbf{Instruction Words} \\
        \midrule
        \texttt{LLaMa-2-7b} & Zero-shot-CoT & 24.26\%  & \textit{Let's think step by step} \\
         & Few-shot-CoT & 24.87\%  & \textit{two exemplars + Let's think step by step} \\
         & \textbf{OPRO} & \textbf{29.81\%} & \textbf{\textit{The correlation is present}} \\
        \midrule
        \texttt{Mistral 7B} & Zero-shot-CoT & 37.52\%  & \textit{Let's think step by step} \\
         & Few-shot-CoT & \textbf{38.13\%}  & \textbf{\textit{two exemplars + Let's think step by step}} \\
         & \textbf{OPRO} & 32.13\% & \textit{Using the provided information, we can find the solution} \\
        \midrule
        \texttt{LLaMa-2-13b} & Zero-shot-CoT & 32.75\% & \textit{Let's think step by step} \\
         & \textbf{Few-shot-CoT} & \textbf{37.15\%} & \textbf{\textit{two exemplars + Let's think step by step}} \\
         & OPRO & 31.24\%  & \textit{Let's think about} \\
        \midrule
        \texttt{LLaMa-2-70b} & Zero-shot-CoT & 39.35\% & \textit{Let's think step by step} \\
         & \textbf{Few-shot-CoT} & \textbf{48.67\%} & \textbf{\textit{two exemplars + Let's think step by step}} \\
         & OPRO & 27.98\% & \textit{The correlation is present} \\
        \midrule
        \texttt{Gemini-Pro} & Zero-shot-CoT & 71.29\% & \textit{Let's think step by step} \\
         & Few-shot-CoT & 69.67\% & \textit{two exemplars + Let's think step by step} \\
         & \textbf{OPRO} & \textbf{76.92\%} & \textbf{\textit{To attain the utmost precision in solving diverse grade school mathematical problems, meticulously adhere to this comprehensive and rigorously developed methodology:}} \\
        \bottomrule
    \end{tabular}
    \vspace{-2mm}
\end{table*}

In this section, we aim to replicate the OPRO framework with small-scale LLMs to assess its efficacy in identifying optimal instruction words. The instruction position is added to the beginning of the LLM output.
\subsection{Experiment setup}
\textbf{Datasets and Models.} We selected models from two distinct categories: small-scale and large-scale. Within the small-scale category, we focused on the Llama family, evaluating \texttt{LLaMa-2-7b}, \texttt{LLaMa-2-13b}, and \texttt{LLaMa-2-70b}. We also conduct experiments with \texttt{Mistral 7B}~\cite{Jiang2023Mistral7} to test the generalizability of the findings. For insights into large-scale LLM performance, we conducted parallel experiments on \texttt{Gemini-Pro}~\cite{Anil2023GeminiAF}.
Following the OPRO paper, all experiments in this paper are conducted with GSM8K, a benchmark of grade school math word problems, with 7,373 training samples and 1,319 test samples.

\textbf{Baselines and Implementations.} We focus on three well-adapted prompting designs in the experiments, including Zero-shot-CoT~\cite{kojima-2022-zeroshotreasoners}, Few-shot-CoT~\cite{wei-2022-chainofthought}, and OPRO~~\cite{yang-2023-llmasoptimizer}. We rigorously follow the original OPRO paper~\cite{yang-2023-llmasoptimizer} for the implementation details. Specifically, we only use the same model architectures for the optimizer and scorer in the main experiment, but these are two independent LLMs. More details about the implementations are shown in the Appendix.

%
%
\subsection{Main Results}
\label{sec:main_results}
We evaluated various prompting strategies across different LLM scales, detailed in Table~\ref{tab:eval_result}, maintaining consistent model architectures for both optimizer and scorer. 
The \texttt{Gemini-Pro} model demonstrates OPRO's effectiveness, notably surpassing CoT baselines, in line with previous findings~\cite{yang-2023-llmasoptimizer}. This underscores OPRO's advantage with large-scale LLMs in optimizing task performance.

Conversely, OPRO's results with \texttt{Mistral 7B}, \texttt{LLaMa-2-13B}, and \texttt{LLaMa-2-70B} fall short of Zero-shot-CoT and Few-shot-CoT benchmarks, revealing these models' limitations in optimization and their inability to outperform basic \textit{"Let's think step by step"} prompts. Notably, the highest performance is observed with Few-shot-CoT, suggesting that for small-scale LLMs, direct instructions providing clear guidance on both the objectives and methodologies are most effective. This aligns with earlier discussions in Section~\ref{sec:mot}, highlighting the insufficient self-optimization capabilities of smaller-scale LLMs in generating optimal instruction words. The results with \texttt{Mistral 7B} validate our argument among different model architectures.

\begin{figure*}[t]
    \centering
    \begin{subfigure}{0.32\textwidth}
        \centering
        \includegraphics[width=\linewidth]{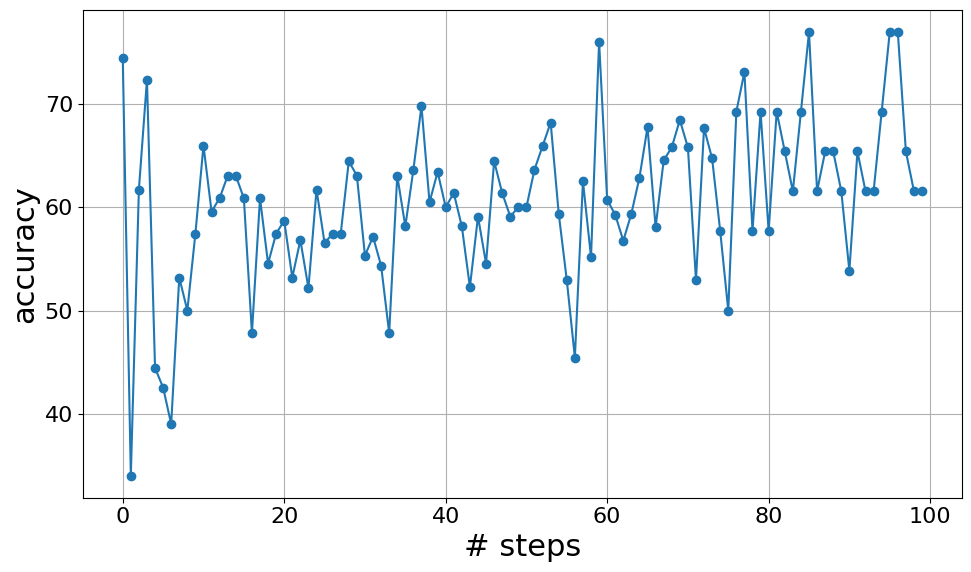}
        \caption{\texttt{Gemini-Pro} optimizer (scorer: \texttt{Gemini-Pro})}
        \label{fig:gemini_gemini}
    \end{subfigure}%
    \hfill
    \begin{subfigure}{0.32\textwidth}
        \centering
        \includegraphics[width=\linewidth]{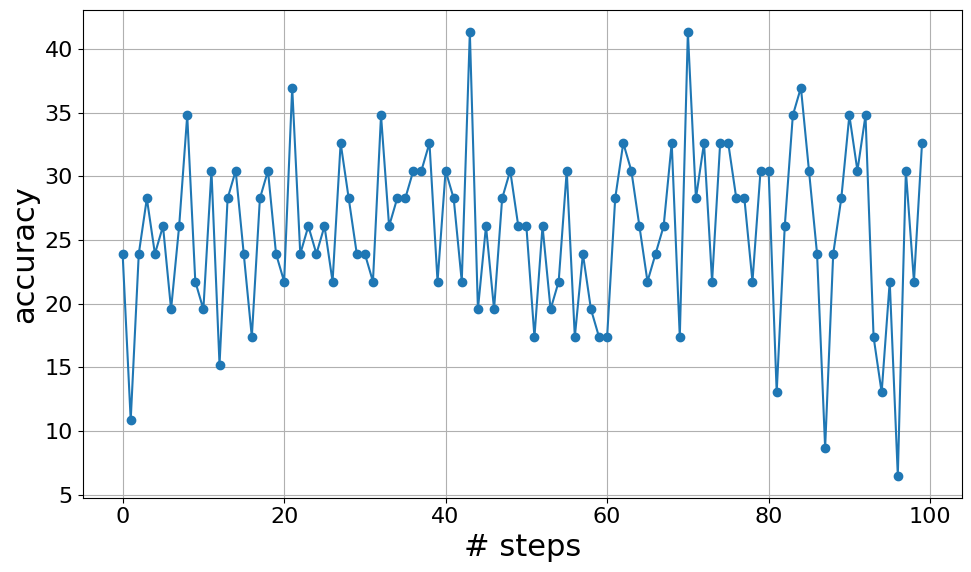}
        \caption{\texttt{LLaMa-2-13b} optimizer (scorer: \texttt{Gemini-Pro})}
        \label{fig:gemini_13b}
    \end{subfigure}%
    \hfill
    \begin{subfigure}{0.32\textwidth}
        \centering
        \includegraphics[width=\linewidth]{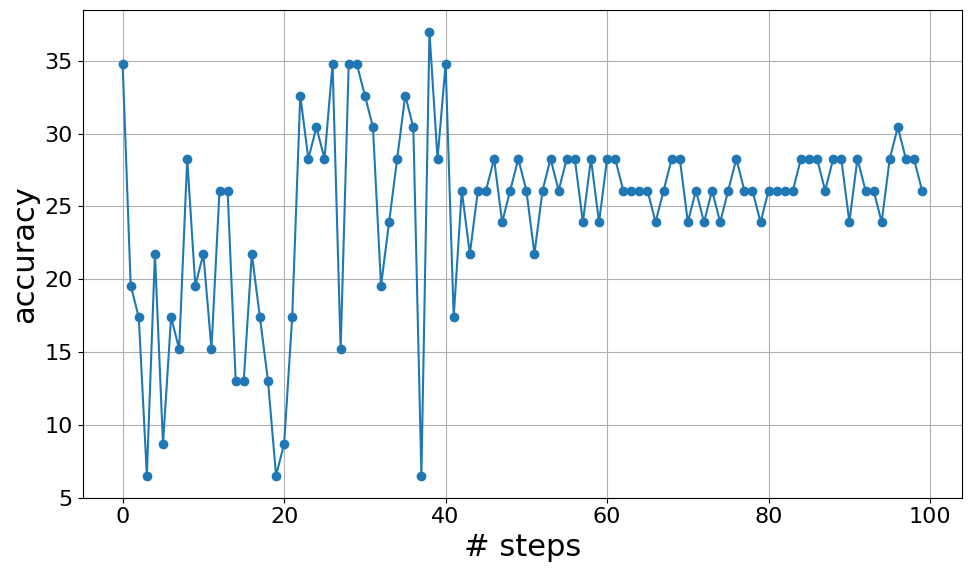}
        \caption{\texttt{LLaMa-2-13b} optimizer (scorer: \texttt{LLaMa-2-13b})}
        \label{fig:13b_13b}
    \end{subfigure}
    \caption{Prompt optimization curve on GSM8K using \texttt{Gemini-Pro} and \texttt{LLaMa-2-13b}.}
    \label{fig:overall}
\end{figure*}

\textbf{Analysis of Generated Instruction Words.} A closer examination of OPRO's instruction generation reveals significant insights into its optimization efficacy. In \texttt{LLaMa-2-13B}, the instructions generated by OPRO resemble the traditional \textit{"Let's think step by step"} prompt, showcasing some optimization capacity but failing to yield the optimal solution. This scenario underscores the inadequate self-optimization skills of smaller-scale LLMs, contrasting sharply with OPRO's performance in \texttt{Gemini-Pro}. For \texttt{Gemini-Pro}, OPRO crafts instructions that aptly include \textit{"grade school mathematical problems"}, indicating superior optimization and understanding that aligns closely with the task. The disparity in output between the smaller and larger-scale models corroborates the preliminary hypothesis: OPRO's optimization approach falls short in smaller-scale LLMs due to their limited self-optimization abilities.

\section{Limitations of Self-Optimization Prompting in Small-Scale LLMs}
\label{sec:discuss}

\textbf{Small-scale LLMs could not support self-optimization.} 
Our analysis, presented in Table~\ref{tab:eval_result}, assesses how small-scale LLMs fare when serving dual roles in optimization and scoring.
Further, Figure~\ref{fig:overall} illustrates the prompt optimization trajectory for \texttt{LLaMa-2-13b} and \texttt{Gemini-Pro}. OPRO's efficacy in large-scale LLMs like \texttt{Gemini-Pro} (Figure~\ref{fig:gemini_gemini}), consistent with previous studies~\cite{yang-2023-llmasoptimizer}.
Notably, transitioning the scorer from \texttt{LLaMa-2-13b} to \texttt{Gemini-Pro}, while maintaining \texttt{LLaMa-2-13b} as the optimizer, yields a 5\% accuracy increase (Figures~\ref{fig:13b_13b} and~\ref{fig:gemini_13b}).
This highlights \texttt{LLaMa-2-13b}'s inadequacy as a scorer to formulate effective optimization targets, thereby constraining optimal solution discovery.

This finding is in line with recommendations from existing literature~\cite{Hsieh2023DistillingSO}, where leveraging outputs from larger LLMs to enhance smaller models reflects our experimental observations. Furthermore, recent literature indicates that without additional inputs, LLMs struggle to self-improve~\cite{Huang2023SelfCorrect}. Interestingly, upgrading the scorer model only minimally affects performance, implying the optimizer may not fully leverage the advanced capabilities of a superior scorer in OPRO's context, leading to suboptimal prompt generation. As a result, due to the limited inference ability, small-scale LLMs could not support self-optimization for prompting paradigms.

\textbf{Human-Crafted Elements and Their Impacts.}
OPRO aims to automate instruction word discovery, minimizing human intervention through LLM capabilities. Yet, our findings indicate significant variability in performance tied to manually designed meta-instructions within OPRO, especially in small-scale LLMs.
We evaluated four distinct meta-instruction texts as shown in Table~\ref{tab:meta_prompt} in the Appendix with \texttt{LLaMa-2-13b}, with results detailed in Table~\ref{tab:meta_ins_comp}. Huge variance on accuracy underscores the critical influence of human-crafted elements on OPRO performance. Despite OPRO's goal of streamlining prompt optimization, it remains reliant on human-crafted meta-instructions, the same as the traditional Zero-shot-CoT approaches. This reliance is echoed in previous research \cite{Zhou2023RevisitAP}, which found that manual prompting typically surpasses automated approaches, a conclusion consistent with our observations in Table~\ref{tab:eval_result}.

\begin{table}[h]
    \caption{OPRO evaluation performance with different meta instructions using \texttt{LLaMa-2-13b} as optimizer. The detailed texts are shown in Table~\ref{tab:meta_prompt} in Appendix.}
    \centering
    \small
    \begin{tabular}{c c m{2.8cm}}
        \toprule
        \textbf{Meta Instruction} & \textbf{Accuracy} & \textbf{Instruction Words}\\
        \midrule
        Text 1 & 17.59\% & \textit{Congratulations! You're a math genius!}\\
        \addlinespace
        Text 2 & 10.39\% & \textit{Now, let's try another problem:}\\
        \addlinespace
        Text 3 & 22.82\% & \textit{The precise answer is}\\
        \addlinespace
        Text 4 & 31.24\% & \textit{Let’s think about}\\
        \bottomrule
    \end{tabular}
    \label{tab:meta_ins_comp}
\end{table}

\begin{table}[h]
    \caption{Approximate input and output tokens with \texttt{Gemini Pro} until optimal instruction words was reached, and approximate computation time in hours.}
    \centering
    \small
    \begin{tabular}{l c c c}
        \toprule
        \textbf{} & \textbf{Zero-shot-CoT} & \textbf{Few-shot-CoT} & \textbf{OPRO}\\
        \midrule
        Input & 6 & 130 & 96,289\\
        Output & 0 & 0 & 170,448\\
        Time (hrs) & 4 & 5 & 21\\
        \bottomrule
    \end{tabular}
    \label{tab:token}
\end{table}

\textbf{Analysis of System Efficiency.}
Recent automatic prompt works~\cite{Fernando2023PromptbreederSS, yang-2023-llmasoptimizer,Ma2024AreLL} have largely overlooked system efficiency for searching instructions. 
In Table~\ref{tab:token}, we examine the efficiency of using \texttt{Gemini Pro} API across three methodologies by comparing input and output tokens and computational time required to achieve the accuracies listed in Table~\ref{tab:eval_result}. 
The token counts are based on a \textit{word-based tokenization} approach.
OPRO incurs a notably higher token count, attributed to the scorer's evaluation process in each meta-prompt generation cycle. Additionally, OPRO's computational time far exceeds that of alternative methods. These results suggest that the efficiency trade-offs associated with OPRO, given its extensive computational demands, may not align with the marginal performance enhancements it offers.
\section{Conclusion}
With empirical results, we demonstrate that small-scale LLMs are limited in self-optimization capacity, which causes OPRO is not effective for small-scale LLMs.
In addition, our findings underscore OPRO's dependency on scorer performance and manually designed prompts, despite the effort to automate the process. We suggest the future automatic prompting engineering consider both model capabilities and system efficiencies.

\textbf{Limitation and Future Study.} Our study's scope was limited by computational resources, excluding other self-optimization strategies like EvoPrompt and APO due to their extensive prompt generation time.
Our future research will extend to enhancing the interpretability and depth of error analysis, alternative optimization metrics, bias considerations, or hyperparameter tuning impacts based on our current findings.

\section{Acknowledgement}
We thank the reviewers for their helpful comments. This work is in part supported by research gifts from the USC Amazon Center for Secure and Trusted AI and Intel. This work is supported in part by Samsung Electronics America.

\bibliography{custom}

\begin{thebibliography}{20}
\expandafter\ifx\csname natexlab\endcsname\relax\def\natexlab#1{#1}\fi

\bibitem[{Fernando et~al.(2023)Fernando, Banarse, Michalewski, Osindero, and Rockt{\"a}schel}]{Fernando2023PromptbreederSS}
Chrisantha Fernando, Dylan~S. Banarse, Henryk Michalewski, Simon Osindero, and Tim Rockt{\"a}schel. 2023.
\newblock \href {https://api.semanticscholar.org/CorpusID:263310323} {Promptbreeder: Self-referential self-improvement via prompt evolution}.
\newblock \emph{ArXiv}, abs/2309.16797.

\bibitem[{Gemini~Team(2023)}]{Anil2023GeminiAF}
Google Gemini~Team. 2023.
\newblock \href {https://api.semanticscholar.org/CorpusID:266361876} {Gemini: A family of highly capable multimodal models}.
\newblock \emph{ArXiv}, abs/2312.11805.

\bibitem[{Guo et~al.(2023)Guo, Wang, Guo, Li, Song, Tan, Liu, Bian, Yang, University, and Research}]{Guo2023ConnectingLL}
Qingyan Guo, Rui Wang, Junliang Guo, Bei Li, Kaitao Song, Xu~Tan, Guoqing Liu, Jiang Bian, Yujiu Yang, Tsinghua University, and Microsoft Research. 2023.
\newblock \href {https://api.semanticscholar.org/CorpusID:262012566} {Connecting large language models with evolutionary algorithms yields powerful prompt optimizers}.
\newblock \emph{ArXiv}, abs/2309.08532.

\bibitem[{Hsieh et~al.(2023)Hsieh, Li, Yeh, Nakhost, Fujii, Ratner, Krishna, Lee, and Pfister}]{Hsieh2023DistillingSO}
Cheng-Yu Hsieh, Chun-Liang Li, Chih-Kuan Yeh, Hootan Nakhost, Yasuhisa Fujii, Alexander~J. Ratner, Ranjay Krishna, Chen-Yu Lee, and Tomas Pfister. 2023.
\newblock \href {https://api.semanticscholar.org/CorpusID:258461606} {Distilling step-by-step! outperforming larger language models with less training data and smaller model sizes}.
\newblock \emph{ArXiv}, abs/2305.02301.

\bibitem[{Huang et~al.(2023)Huang, Chen, Mishra, Zheng, Yu, Song, and Zhou}]{Huang2023SelfCorrect}
Jie Huang, Xinyun Chen, Swaroop Mishra, Huaixiu~Steven Zheng, Adams~Wei Yu, Xinying Song, and Denny Zhou. 2023.
\newblock Large language models cannot self-correct reasoning yet.
\newblock \emph{arXiv:2310.01798}.

\bibitem[{Jiang et~al.(2023)Jiang, Sablayrolles, Mensch, Bamford, Chaplot, de~Las~Casas, Bressand, Lengyel, Lample, Saulnier, Lavaud, Lachaux, Stock, Scao, Lavril, Wang, Lacroix, and Sayed}]{Jiang2023Mistral7}
Albert~Qiaochu Jiang, Alexandre Sablayrolles, Arthur Mensch, Chris Bamford, Devendra~Singh Chaplot, Diego de~Las~Casas, Florian Bressand, Gianna Lengyel, Guillaume Lample, Lucile Saulnier, L'elio~Renard Lavaud, Marie-Anne Lachaux, Pierre Stock, Teven~Le Scao, Thibaut Lavril, Thomas Wang, Timoth{\'e}e Lacroix, and William~El Sayed. 2023.
\newblock \href {https://api.semanticscholar.org/CorpusID:263830494} {Mistral 7b}.
\newblock \emph{ArXiv}, abs/2310.06825.

\bibitem[{Kojima et~al.(2022)Kojima, Gu, Reid, Matsuo, and Iwasawa}]{kojima-2022-zeroshotreasoners}
Takeshi Kojima, Shixiang~Shane Gu, Machel Reid, Yutaka Matsuo, and Yusuke Iwasawa. 2022.
\newblock Large language models are zero-shot reasoners.
\newblock \emph{Advances in Neural Information Processing Systems}, 35:22199--22213.

\bibitem[{Liu et~al.(2021)Liu, Zheng, Du, Ding, Qian, Yang, and Tang}]{Liu2021GPTUT}
Xiao Liu, Yanan Zheng, Zhengxiao Du, Ming Ding, Yujie Qian, Zhilin Yang, and Jie Tang. 2021.
\newblock \href {https://api.semanticscholar.org/CorpusID:232269696} {Gpt understands, too}.
\newblock \emph{ArXiv}, abs/2103.10385.

\bibitem[{Ma et~al.(2024)Ma, Wang, Zhou, Li, Du, Gui, Zhang, and Huang}]{Ma2024AreLL}
Ruotian Ma, Xiaolei Wang, Xin Zhou, Jian Li, Nan Du, Tao Gui, Qi~Zhang, and Xuanjing Huang. 2024.
\newblock \href {https://api.semanticscholar.org/CorpusID:267413214} {Are large language models good prompt optimizers?}
\newblock \emph{ArXiv}, abs/2402.02101.

\bibitem[{OpenAI(2020)}]{brown-2020-fewshotlearners}
OpenAI. 2020.
\newblock Language models are few-shot learners.
\newblock \emph{NeurIPS}.

\bibitem[{Paszke et~al.(2019)Paszke, Gross, Massa, Lerer, Bradbury, Chanan, Killeen, Lin, Gimelshein, Antiga, Desmasion, Kopf, Yang, DeVito, Raison, Tejani, Chilamkurthy, Steiner, Fang, Bai, and Chintala}]{pytorch}
Adam Paszke, Sam Gross, Francisco Massa, Adam Lerer, James Bradbury, Gregory Chanan, Trevor Killeen, Zeming Lin, Natalia Gimelshein, Luca Antiga, Alban Desmasion, Andreas Kopf, Edward Yang, Zachary DeVito, Martin Raison, Alykhan Tejani, Sasank Chilamkurthy, Benoit Steiner, Lu~Fang, Junjie Bai, and Soumith Chintala. 2019.
\newblock Pytorch: An imperative style, high performance deep learning library.
\newblock In \emph{Advances in Neural Information Processing Systems 32}, pages 8024--8035.

\bibitem[{Pryzant et~al.(2023)Pryzant, Iter, Li, Lee, Zhu, and Zeng}]{Pryzant2023AutomaticPO}
Reid Pryzant, Dan Iter, Jerry Li, Yin~Tat Lee, Chenguang Zhu, and Michael Zeng. 2023.
\newblock \href {https://api.semanticscholar.org/CorpusID:258546785} {Automatic prompt optimization with "gradient descent" and beam search}.
\newblock In \emph{Conference on Empirical Methods in Natural Language Processing}.

\bibitem[{Touvron et~al.(2023)Touvron, Martin, Stone, Albert et~al.}]{Touvron2023Llama2O}
Hugo Touvron, Louis Martin, Kevin~R. Stone, Peter Albert, et~al. 2023.
\newblock \href {https://api.semanticscholar.org/CorpusID:259950998} {Llama 2: Open foundation and fine-tuned chat models}.
\newblock \emph{ArXiv}, abs/2307.09288.

\bibitem[{Wan et~al.(2023)Wan, Wang, Liu, Alam, Zheng, Liu, Qu, Yan, Zhu, Zhang, Chowdhury, and Zhang}]{Wan2023EfficientLL}
Zhongwei Wan, Xin Wang, Che Liu, Samiul Alam, Yu~Zheng, Jiachen Liu, Zhongnan Qu, Shen Yan, Yi~Zhu, Quanlu Zhang, Mosharaf Chowdhury, and Mi~Zhang. 2023.
\newblock \href {https://api.semanticscholar.org/CorpusID:266044196} {Efficient large language models: A survey}.
\newblock \emph{ArXiv}, abs/2312.03863.

\bibitem[{Wang et~al.(2022)Wang, Wei, Schuurmans, Le, hsin Chi, and Zhou}]{Wang2022SelfConsistencyIC}
Xuezhi Wang, Jason Wei, Dale Schuurmans, Quoc Le, Ed~Huai hsin Chi, and Denny Zhou. 2022.
\newblock \href {https://api.semanticscholar.org/CorpusID:247595263} {Self-consistency improves chain of thought reasoning in language models}.
\newblock \emph{ArXiv}, abs/2203.11171.

\bibitem[{Wei et~al.(2022)Wei, Wang, Schuurmans, Bosma, Chi, Le, and Zhou}]{wei-2022-chainofthought}
Jason Wei, Xuezhi Wang, Dale Schuurmans, Maarten Bosma, Ed~Chi, Quoc Le, and Denny Zhou. 2022.
\newblock Chain of thought prompting elicits reasoning in large language models.
\newblock \emph{arXiv:2201.11903}.

\bibitem[{Yang et~al.(2023)Yang, Wang, Lu, Liu, Le, Zhou, and Chen}]{yang-2023-llmasoptimizer}
Chengrun Yang, Xuezhi Wang, Yifeng Lu, Hanxiao Liu, Quoc~V Le, Denny Zhou, and Xinyun Chen. 2023.
\newblock Large language models as optimizers.
\newblock \emph{arXiv preprint arXiv:2309.03409}.

\bibitem[{Yao et~al.(2023)Yao, Yu, Zhao, Shafran, Griffiths, Cao, and Narasimhan}]{Yao2023TreeOT}
Shunyu Yao, Dian Yu, Jeffrey Zhao, Izhak Shafran, Thomas~L. Griffiths, Yuan Cao, and Karthik Narasimhan. 2023.
\newblock \href {https://api.semanticscholar.org/CorpusID:258762525} {Tree of thoughts: Deliberate problem solving with large language models}.
\newblock \emph{ArXiv}, abs/2305.10601.

\bibitem[{Zhou et~al.(2022)Zhou, Muresanu, Han, Paster, Pitis, Chan, and Ba}]{Zhou2022LargeLM}
Yongchao Zhou, Andrei~Ioan Muresanu, Ziwen Han, Keiran Paster, Silviu Pitis, Harris Chan, and Jimmy Ba. 2022.
\newblock \href {https://api.semanticscholar.org/CorpusID:253265328} {Large language models are human-level prompt engineers}.
\newblock \emph{ArXiv}, abs/2211.01910.

\bibitem[{Zhou et~al.(2023)Zhou, Zhao, Shumailov, Mullins, and Gal}]{Zhou2023RevisitAP}
Yulin Zhou, Yiren Zhao, Ilia Shumailov, Robert Mullins, and Yarin Gal. 2023.
\newblock Revisiting automated prompting: Are we actually doing better?
\newblock In \emph{The 61st Annual Meeting of the Association of Computational Linguistics}.

\end{thebibliography}
\bibliographystyle{acl_natbib}

\appendix
\newpage
\onecolumn
\section{Experimental Details}

\subsection{Models and Test Environment}
We implemented the experiments using PyTorch~\cite{pytorch}, and conducted our experiments on two NVIDIA A100 GPUs. We tested \texttt{LLaMa-2-7b}, \texttt{LLaMa-2-13b}, \texttt{LLaMa-2-70b}, and \texttt{Gemini-Pro} in the experiments. We downloaded LLaMa models from Hugging Face and tested them locally on GPUs. For \texttt{Gemini-Pro}, we referenced the model via the Gemini API. The links for the models are shown below.
\vspace{0.5cm}

\texttt{LLaMa-2-7b} link:

\href{https://huggingface.co/meta-llama/Llama-2-7b-chat-hf}{https://huggingface.co/meta-llama/Llama-2-7b-chat-hf}
\vspace{0.5cm}

\texttt{LLaMa-2-13b} link:

\href{https://huggingface.co/meta-llama/Llama-2-13b-chat-hf}{https://huggingface.co/meta-llama/Llama-2-13b-chat-hf}

\vspace{0.5cm}

\texttt{LLaMa-2-70b} link:

\href{https://huggingface.co/meta-llama/Llama-2-70b-chat-hf}{https://huggingface.co/meta-llama/Llama-2-70b-chat-hf}

\vspace{0.5cm}

\texttt{Gemini-Pro} link:

\href{https://ai.google.dev/models/gemini}{https://ai.google.dev/models/gemini}
\vspace{0.5cm}

\subsection{Prompting Methods}
\begin{enumerate}
    \item Zero-shot-CoT: The zero-shot instruction \textit{"Let's think step by step"}~\cite{kojima-2022-zeroshotreasoners} would be added before each answers.
    \item Few-shot-CoT: We randomly select two samples with procedures~\cite{wei-2022-chainofthought} from the training set serving as the problem description before the test question.
    \item OPRO: We rigorously follow the original paper~\cite{yang-2023-llmasoptimizer} for the implementation details. Our experiment utilized a meta-prompt, as illustrated in Figure~\ref{fig:prompt}, with the optimization process spanning 100 iterations. In each iteration, we sampled 3.5\% of GSM8K training examples as a validation set for scorer LLM. We used the meta-prompt to generate eight new instructions with the optimizer LLM, updating the trajectory with these instructions and their scores in each interaction. The meta-prompt included the top 20 instructions and three random training exemplars.
\end{enumerate}
\vspace{0.5cm}

\newpage
\section{Meta-Prompt Design}
\begin{figure}[h]
\centering
\includegraphics[width=0.6\linewidth]{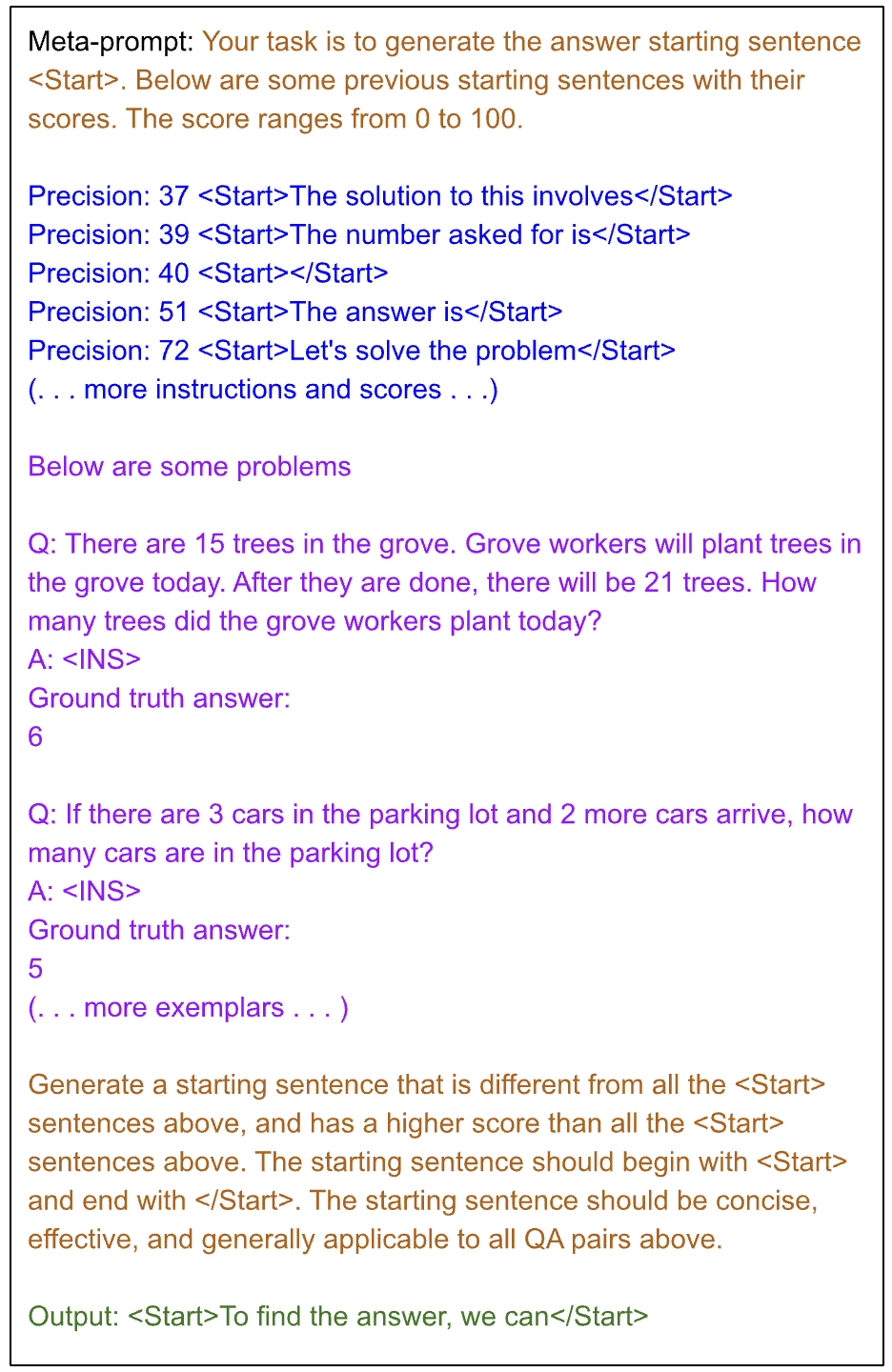}
\caption{An example of the meta-prompt and its output for GSM8K dataset. 
The \textcolor{brown}{orange} text are meta-instructions; 
the \textcolor{blue}{blue} text are solution-score pairs; 
the \textcolor{violet}{purple} text are exemplar questions and optimization targets; 
the \textcolor{green}{green} text are output by LLM.
}
\label{fig:prompt}
\end{figure}

Figure \ref{fig:prompt} shows an example of the meta-prompt used in our implementation of OPRO. We rigorously followed the original open source code provided by Google Deep Mind (\href{https://github.com/google-deepmind/opro}{https://github.com/google-deepmind/opro}). The two example problems are exemplars randomly selected from the training set of GSM8K to support the meta-prompt as the problem description.

\newpage
\section{Meta-Instruction Design}
To test the robustness of meta-prompt, we experimented with several slightly adjusted meta-instructions. \textit{Text 4} strictly follows Yang et al.'s design \cite{yang-2023-llmasoptimizer}. To prevent human invention on the prompt design, we input \textit{Text 4} into ChatGPT (\href{https://openai.com/chatgpt}{https://openai.com/chatgpt}) to generate the three other prompts. Precision scores are replaced with the scores produced by the scorer in the later evaluation steps during the computation of OPRO. Table \ref{tab:meta_ins_comp} shows the results of the different meta-instructions on performances.

\begin{table*}[ht]
    \centering
    \caption{Meta instructions used in OPRO}
    \begin{tabular}{l m{13cm}}
        \toprule
        \textbf{Meta Instruction} & \textbf{Text}\\
        \midrule
        Text 1 & Create a new piece of text as an instruction at the beginning of the answer to enhance the precision in solving diverse grade school math problems. We want the precision of the text to be higher as possible. Range of Precision is 0 to 100. For example, Precision: 4 <Text>A dime</Text>, Precision: 17 <Text>The answer is a function. It is</Text>.\\
        \addlinespace
        Text 2 & Write a new text for instruction use before the answer in the Q\&A pair to help solving the grade school math problems. We want to precision of the text to be as high as possible, ranging from 0 to 100. For example, Precision: 4 <Text>A quarter</Text>, Precision: 25 <Text>Now find the answer.</Text>.\\
        \addlinespace
        Text 3 & Create a line of instruction, with precision 0 to 100. The text will be placed at the start of the answer, to assist in solving grade school mathematical problems. Some example text and score pairs are: Precision: 29 <Text>The numeric answer to this question is:</Text>\\
        \addlinespace
        Text 4 & Your task is to generate the answer starting sentence <Start>. Below are some previous starting sentences with their scores. The score ranges from 0 to 100. Precision: 37 <Start>The solution to this involves</Start>, Precision: 39 <Start>The number asked for is</Start>\\
        \bottomrule
    \end{tabular}
    \label{tab:meta_prompt}
\end{table*}


\end{document}